\documentclass[letterpaper]{jpsj3}
\usepackage{txfonts}

\usepackage{color}
\usepackage{tikz}
\usepackage{bm}

\newcommand\fig[1]{Fig.~\ref{#1}}
\newcommand\eq[1]{Eq.~(\ref{#1})}
\newcommand{\R}{\mathbb{R}}
\newcommand{\lt}{<}
\renewcommand\vec[1]{\bm{#1}}
\DeclareMathOperator*{\argmin}{arg~min}
\DeclareMathOperator*{\argmax}{arg~max}

\title{
Random Postprocessing for Combinatorial Bayesian Optimization
}

\author{Keisuke Morita$^1$, Yoshihiko Nishikawa$^{1,2}$, and Masayuki Ohzeki$^{1,3,4,5}$}
\inst{
$^1$Graduate School of Information Sciences, Tohoku University, Miyagi, 980-8564, Japan \\
$^2$Department of Physics, Kitasato University, Kanagawa, 252-0373, Japan\\
$^3$Department of Physics, Tokyo Institute of Technology, Tokyo, 152-8551, Japan \\
$^4$International Research Frontier Initiative, Tokyo Institute of Technology, Tokyo, 108-0023, Japan \\
$^5$Sigma-i Co., Ltd., Tokyo, 108-0075, Japan
}

\abst{
Model-based sequential approaches to discrete ``black-box'' optimization, including Bayesian optimization techniques, often access the same points multiple times for a given objective function in interest, resulting in many steps to find the global optimum.
Here, we numerically study the effect of a postprocessing method on Bayesian optimization that strictly prohibits duplicated samples in the dataset.
We find the postprocessing method significantly reduces the number of sequential steps to find the global optimum, especially when the acquisition function is of \textit{maximum a posteriori} estimation.
Our results provide a simple but general strategy to solve the slow convergence of Bayesian optimization for high-dimensional problems.
}

\begin{document}
\maketitle

Optimizing an expensive-to-evaluate objective function $f: \mathcal X \to \R$ is a common task in various practical situations, such as
crystal structure prediction \cite{yamashita2018crystal},
designing nanostructures \cite{ju2017designing},
automated material discovery \cite{kitai2020designing,homma2020optimization,Tanaka2023},
building electronic circuits \cite{matsumori2022application},
parameter optimization of a quantum circuit \cite{tibaldi2022bayesian},
and finding effective Hamiltonians \cite{tamura2020data,seki2022black}.
Due to the significant expense of evaluating \( f(x) \), it is crucial to identify the globally optimal solution \( x_\text{opt} = \argmin_{x \in \mathcal X} f(x) \) with as few function evaluations as possible.
Bayesian optimization \cite{jones1998efficient,pelikan1999boa,brochu2010tutorial,shahriari2015taking,garnett_bayesoptbook_2023} deals with these difficulties by introducing a statistical model called a \textit{surrogate model} and sequentially estimating $f(x)$.
At each iteration step \( t \), we update the surrogate model \( \hat f \) to fit the already known input-output dataset \( \mathcal D = \{ x^{(i)}, f(x^{(i)})\}_{i=1}^t \).
Using the properties of the updated surrogate model \( \hat f \), we build an \textit{acquisition function} \( \alpha: \mathcal X \to \R \) and optimize it to find the most \textit{promising} input point \( x^{(t+1)} \), namely, \( x^{(t+1)} = \argmin_{x \in \mathcal X} \alpha(x) \).
We then compute the objective function of the chosen input point, \( f(x^{(t+1)}) \) and append the pair to the dataset: \( \mathcal D \gets \mathcal D \cup \{ (x^{(t+1)}, f(x^{(t+1)})) \} \).
This process is repeated until a termination criterion is fulfilled, e.g., exhausting the predetermined maximum number of steps or finding a sample satisfying a desired constraint.

In recent years, several attempts have been made to apply Bayesian optimization to high-dimensional combinatorial optimization problems \cite{baptista2018bayesian,Takahashi2018,oh2019combinatorial,deshwal2021mercer,Koshikawa2021, deshwal2022bayesian,dadkhahi2022fourier,kim2020surrogate,nusslein2023black,Tanaka2023}.
These methods often take a long time to reach the global optimum as the acquisition functions yield the same or nearby points many times \cite{luong2019bayesian} and get stuck in a local optimum.
This could become serious when the next point to compute the objective function is determined from the acquisition function in a deterministic manner \cite{baptista2018bayesian,kadowaki2022lossy}, in which the algorithm can never escape from a local optimum.

Reference~\citen{kitai2020designing} solved the aforementioned issue by randomly selecting a new point as a postprocessing step \cite{kitai2020designing}:
If the next query point \( x^{(t+1)} \) drawn from the acquisition function is already in the dataset \( \mathcal D \), it is rejected, and another randomly chosen point is proposed.
Reference~\citen{kitai2020designing} combined this process with an algorithm using the factorization machines \cite{rendle2010factorization,rendle2012factorization}.
This postprocessing technique is simpler and more straightforward than other methods \cite{papalexopoulos2022constrained} to implement as it does not need any modification to the acquisition function. Our primary interest thus lies in understanding the effects of this postprocessing on Bayesian optimization.

In this letter, we apply the postprocessing method to Bayesian optimization using Thompson sampling, and study its performance for the ground-state search of the Sherrington--Kirkpatrick spin glass model. 
We find that the Bayesian optimization algorithm with MAP estimation and random postprocessing outperforms its Thompson sampling variant.
Our results imply that random postprocessing can improve Bayesian optimization for high-dimensional problems.

We focus on the ground-state search of the Sherrington--Kirkpatrick (SK) model \cite{sherrington1975solvable} defined by the Hamiltonian
\begin{align}
H
&= \frac{1}{\sqrt N} \sum_{i \lt j} J_{ij} s_i s_j \\
&= \frac{1}{\sqrt N} \sum_{i \lt j} J_{ij} (2 x_i - 1) (2 x_j - 1).
\label{eq:SK_x}
\end{align}
Here, \( N \) is the number of spins, \( s_i \in \{-1, 1\} \), \( x_i = (s_i + 1) / 2 \in \{0, 1\} \), and the interaction between spin $i$ and $j$ is drawn from the normal distribution with zero mean and variance $J^2$, i.e., \( J_{ij} \sim \mathcal N(0, J^2)\).
We use a second-order surrogate model as in the `Bayesian optimization of combinatorial structures' (BOCS) algorithm \cite{baptista2018bayesian} for simplicity, which is given by
\begin{align}
    \hat f_{\vec w}(x)
    = w_0 + \sum_i w_i x_i + \sum_{i \lt j} w_{ij} x_i x_j = \vec w^\top \vec z,
\end{align}
where \( x \in \{0, 1\}^N \) is the input vector and \( w_{i}, w_{ij} \in \R \) are parameters of the model.
This model is linear with respect to \( \vec w = (w_0, w_1, \dots, w_N, w_{12}, \dots, w_{(N-1)N})^\top \in \R^P \) and \( \vec z = (1, x_1, \dots, x_N, x_1 x_2, \dots, x_{N-1} x_N)^\top \in \{0, 1\}^P \) where \( P=1 + N + \binom{N}{2} \).
We conduct linear regression to estimate the model parameters \( \vec w \) from the dataset \( \mathcal D = \{(\vec x^{(i)}, H(\vec x^{(i)}))\} \) as follows. 
We first normalize the observed energies $\{ H(\vec x^{(i)}) \}_{i=0,1,\cdots,|\mathcal D| - 1}$ as
\begin{align}
y^{(i)} = 2 \frac{
    H(\vec x^{(i)}) - \min_{j \in \mathcal D(t)} H(\vec x^{(j)})
}{
    \max_{j \in \mathcal D(t)} H(\vec x^{(j)}) - \min_{j \in \mathcal D(t)} H(\vec x^{(j)} )
} - 1.
\label{eq:data_normalization}
\end{align}
so that $y^{(i)} \in [-1, 1]$. 
While our algorithm works without this normalization, it slightly shortens the time to find the ground state.
We then assume that $y^{(i)}$ is distributed according to the normal distribution with variance \( \sigma_y^2 \):
\begin{align}
y^{(i)} | \vec x^{(i)}, \vec w \sim \mathcal N(\vec w^\top \vec z^{(i)}, \sigma_y^2).
\end{align}
Whereas the original BOCS algorithm uses the horseshoe prior distribution \cite{carvalho2009handling,carvalho2010horseshoe,makalic2015simple}, which efficiently works for sparse parametric models \cite{bhadra2019lasso}, we employ the normal prior \( \vec w \sim \mathcal N_t(\vec 0, \sigma_\text{pr}^2 \vec I) \) as the SK model has a dense structure in $\{J_{ij}\}$.
Thanks to the conjugacy of the normal distribution, the posterior distribution is also multivariate normal \( \vec w | \mathcal D \sim \mathcal N_t(\vec m_\text{pos}, \vec  V_\text{pos}) \), with the parameters
\begin{align}
\vec  m_{\rm pos} &= \frac{1}{\sigma^2_y} \vec V_{\rm pos} \bm Z^\top \vec y, \\
\vec V_{\rm pos} &= \sigma_y^2 \left( \bm Z^\top \bm Z + \frac{\sigma^2_{y}}{\sigma^2_{\rm pr}} \vec I \right)^{-1},
\end{align}
where \( \bm Z = (\vec z^{(0)}, \vec z^{(1)}, \dots, \vec z^{(|\mathcal D|-1)})^\top \in \{0, 1\}^{|\mathcal D| \times N} \) and \( \vec y = (y^{(0)}, y^{(1)}, \dots, y^{(|\mathcal D|-1)})^\top \in \R^{|\mathcal D|} \).
The values of hyperparameters are fixed to $\sigma_\text{pr}^2 = J^2$ and $\sigma_y^2 = 10^{-2}J^2$. These values have been found in a preliminary run to optimize our BOCS algorithms.
Hereafter, we refer to BOCS with the normal prior as `nBOCS' following Ref.~\citen{kadowaki2022lossy}.

We adopt Thompson sampling (TS) \cite{thompson1933likelihood,thompson1935criterion,agrawal2013thompson} and MAP estimation to build an acquisition function.
TS draws a sample of the model parameters from the posterior and uses it to construct an acquisition function
\begin{align}
\alpha^\text{TS} = \hat f_{\tilde w}, \quad \text{where } \tilde w \sim p(w | \mathcal D).
\label{eq:TS}
\end{align}
In MAP estimation, on the other hand, we use a set of parameters that maximizes the posterior probability. 
The resultant acquisition function is thus
\begin{align}
\alpha^\text{MAP} = \hat f_{\tilde w}, \quad \text{where } \tilde w = \argmax_{w} p(w | \mathcal D).
\label{eq:MAP}
\end{align}
Note that the acquisition function used in Reference~\citen{kitai2020designing} is also the surrogate model with $\vec w$ optimizing $p(w | \mathcal D)$.
We use simulated annealing (SA) \cite{kirkpatrick1983optimization,vcerny1985thermodynamical,bertsimas1993simulated} to determine the next query point. 
In our annealing schedule for SA, the inverse temperature $\beta(r) = \beta_\text{init} \times (\beta_\text{final} / \beta_\text{init})^{r / r_\text{total}}$ with $r = 0, 1, \cdots, r_\text{total}$ the time step of simulated annealing and $r_\text{total}$ the total number of Monte Carlo sweeps. 
We set $\beta_\text{init} = 10^{-3} / J$ and $r_\text{total} = 10^4$.
We systematically change the final inverse temperature $\beta_\text{final}$, ranging from $10^0/J$ to $10^4/J$, to see how it affects the performance of our algorithm. 
Even though SA is not a deterministic method, it could yield a point already included in the current dataset $\mathcal D$. 
In this case, we could reject it and sample a new one uniformly randomly from $\mathcal X$. 
Once we obtain a sample $\vec x_\text{next}$ not included in $\mathcal D$, we compute $H(\vec x_\text{next})$ and append $(\vec x_\text{next}, H(\vec x_\text{next}))$ to $\mathcal D$. 
This postprocessing strictly prohibits duplicated samples in the dataset $\mathcal D$.

We run nBOCS with the two different acquisition functions, which we refer to as `nBOCS~(TS)' and `nBOCS~(MAP)', respectively, in the following. 
We also study the performances of these algorithms combined with the postprocessing we described above, which we call `nBOCS-Random~(TS)', and `nBOCS-Random~(MAP)', respectively.
Starting with single one randomly chosen pair \( \{(\vec x^{(0)}, H(\vec x^{(0)})\} \) as the initial dataset $\mathcal D(t=0)$, we measure the normalized smallest energy in the dataset $\mathcal D(t) = \{(\vec x^{(i)}, H(\vec x^{(i)}))\}_{i=0,1,\cdots, t-1}$
\begin{align}
[u(t)] = \left[\frac{\min_{j \in \mathcal D(t)}H(\vec x^{(j)}) - H_\text{min}}{H_\text{max} - H_\text{min}}\right],
\label{eq:normalized_energy}
\end{align}
in which the bracket $[\cdot]$ stands for an average over disorder realizations, \(H_\text{min}\) and \(H_\text{max}\) are the minimum and maximum possible energies of each disorder instance, respectively. 
The typical number of disorder instances is $10^2$.
Whereas the ground-state search of the SK model is an NP-hard problem taking an exponentially long time with $N$ in the worst case \cite{arora2005non}, we find the ground state for each instance by the branch-and-bound algorithm \cite{de1995exact,mitchell2002branch} implemented in the Gurobi optimizer \cite{gurobi}.

\begin{figure}[t]
\begin{center}
\includegraphics[width=80mm]{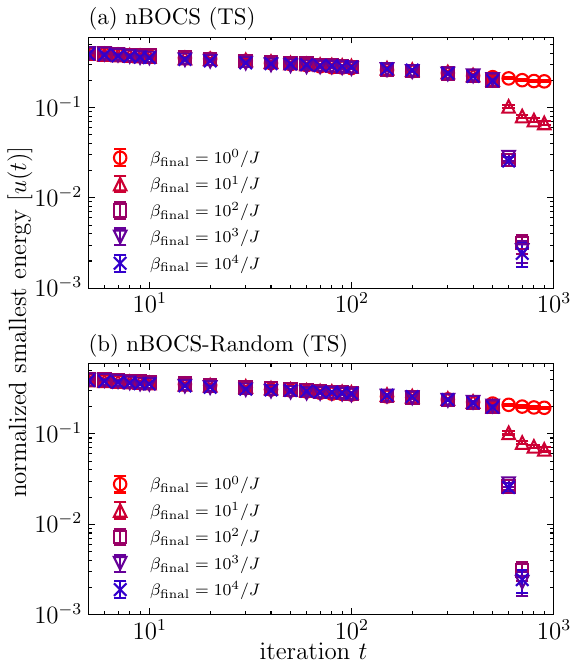}
\caption{ (Color online)
Normalized smallest energy \( [u(t)] \) (\eq{eq:normalized_energy}) as a function of iteration step \( t \) for (a) nBOCS~(TS) and (b) nBOCS-Random~(TS).
The number of spins \( N = 32 \).
}
\label{result:nBOCS-TS}
\end{center}
\end{figure}

\begin{figure}[t]
\begin{center}
\includegraphics[width=80mm]{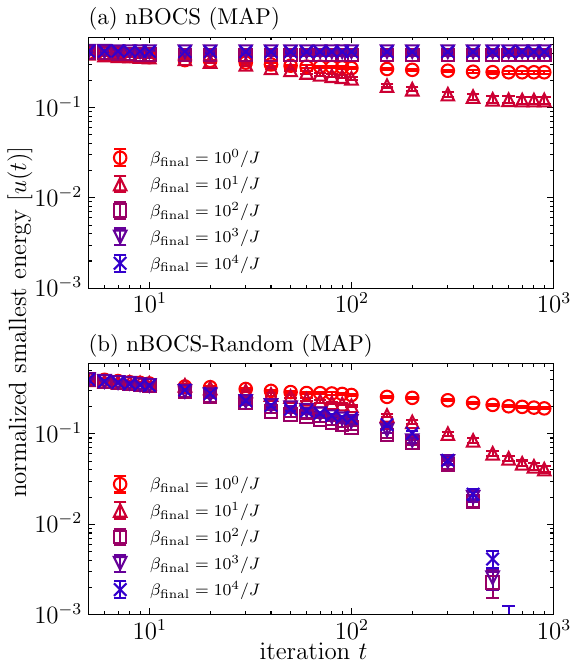}
\caption{ (Color online)
Normalized smallest energy \( [u(t)] \) (\eq{eq:normalized_energy}) as a function of iteration step \( t \) for (a) nBOCS~(MAP) and (b) nBOCS-Random~(MAP).
The number of spins \( N = 32 \).
}
\label{result:nBOCS-MAP}
\end{center}
\end{figure}

Figure \ref{result:nBOCS-TS} shows $[u(t)]$ for nBOCS~(TS) and nBOCS-Random~(TS) with various $\beta_\text{final}$'s. 
The performance strongly depends on $\beta_\text{final}$ and a larger $\beta_\text{final}$ gives smaller iteration steps to find the ground state, but the postprocessing does not change $[u(t)]$ for any $\beta_\text{final}$. 
This comes from the fact that $\alpha^\text{TS}(\vec x)$ (\eq{eq:TS}) rarely yields a point already included in $\mathcal D(t)$ and triggers the postprocessing with a very small probability. Therefore, nBOCS~(TS) and nBOCS-Random~(TS) are virtually identical.
In contrast, the postprocessing improves the performance of nBOCS when using the acquisition function $\alpha^\text{MAP}(\vec x)$ (see \fig{result:nBOCS-MAP}): nBOCS-Random~(MAP) typically finds the ground state within $10^3$ steps if $\beta_\text{final}$ is large enough, whereas $[u(t)]$ of nBOCS~(MAP) with any $\beta_\text{final}$ used in our runs gets stuck at $[u(t)] \gtrsim 10^{-1}$ and is independent of $t$ when $t \gtrsim 5\times 10^2$, indicating that nBOCS~(MAP) cannot find the ground state even in the limit $t \to \infty$.
The parameter $\vec w$ is typically sparse when nBOCS~(MAP) converges to an incorrect one $\vec w_e$: Most components are much smaller than $J$, and only a small fraction have large amplitudes comparable to $J$. Finding the ground state of the surrogate model $\hat f_{\vec w_e}(\vec x)$ with SA is thus easy, and the same point is always appended to $\mathcal D$ even though SA is stochastic, meaning that $\vec w_e$ is a fixed point.
Appending uniformly random samples to $\mathcal D$ drives $\vec w$ away from the incorrect fixed point $\vec w_e$.

\begin{figure}[t]
\begin{center}
\includegraphics[width=80mm]{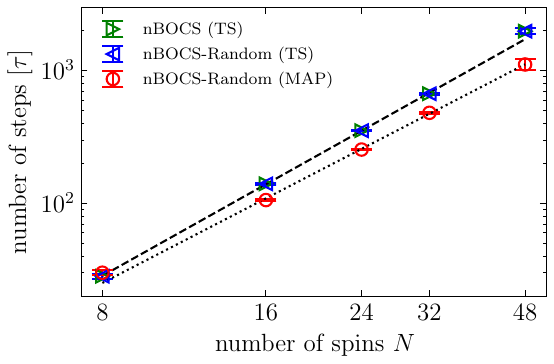}
\caption{ (Color online)
Number of steps $[\tau]$ to reach $[u(t)] = 10^{-3}$ as a function of $N$. 
The broken and dotted curves are power-law growths with exponent $2.3$ and $2.1$, respectively.
}
\label{result:dependence}
\end{center}
\end{figure}

\begin{figure}[t]
\begin{center}
\includegraphics[width=80mm]{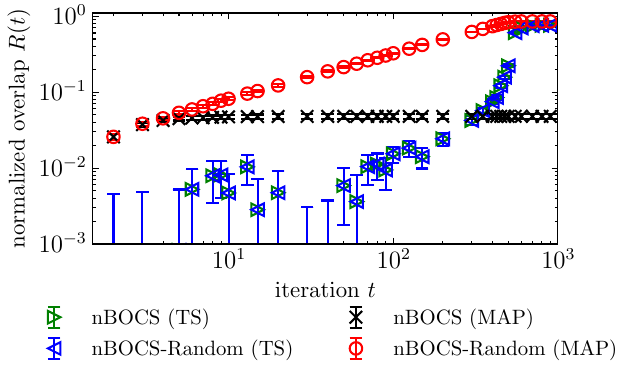}
\caption{ (Color online)
Normalized overlap $R(t)$ (\eq{eq:R}) as a function of iteration step $t$ for nBOCS~(TS), nBOCS-Random~(TS), nBOCS~(MAP), and nBOCS-Random~(MAP). The number of spins $N = 32$. $R(t)$ of nBOCS~(MAP) is completely stuck at a very small value, whereas that of the other algorithms grows with $t$, eventually converging to a large value close to $1$.
}
\label{result:normalized_overlap}
\end{center}
\end{figure}

Now, we characterize the performance of each algorithm by the scaling of the typical number of steps $[\tau]$ needed to find the ground state. 
For each disorder instance, we define $\tau$ as the number of steps where $u(t = \tau) = 10^{-3}$. We have checked that the following results do not change if we set the threshold value to $10^{-6}$. 
In \fig{result:dependence}, we show $[\tau]$ for nBOCS~(TS), nBOCS-Random~(TS), and nBOCS-Random~(MAP) with $\beta_\text{final} = 10^4 / J$, as a function of the number of spins $N$. 
For all the algorithms we have studied, $[\tau]$ grows only algebraically with $N$, with an exponent $z$. 
This indicates that its average-case complexity is much easier than the worst case, as shown in Ref.~\citen{montanari2021optimization}. 
The exponent $z$ depends on algorithms, and nBOCS-Random~(MAP) yields the smallest value, $z = 2.12(3)$, while $z = 2.29(4)$ for the other two algorithms, suggesting that the postprocessing qualitatively changes how the algorithm explores the parameter space. 
This is better seen in the normalized overlap between the parameter of the acquisition function, $\vec w(t)$, and the correct one, $\vec w_J$, determined by \eq{eq:SK_x},
\begin{align}
R(t) = \left[\frac{ \vec w(t)\cdot \vec w_J }{ \|\vec w(t)\| \cdot \|\vec w_J\| }\right].
\label{eq:R}
\end{align}
If $\vec w(t)$ is very close to $\vec w_J$, $R(t) \simeq 1$. We show $R(t)$ for all the algorithms in \fig{result:normalized_overlap}. 
For nBOCS with Thompson sampling, $R(t)$ grows very slowly with $t$ because of random sampling of $\vec w(t) \sim p(\vec w | \mathcal D_t)$. As more samples are appended to $\mathcal D_t$, the covariance matrix $V_\text{pos}$ shrinks, $\vec w(t)$ converges near $\vec w_J$, and $R(t)$ finally becomes large.
On the other hand, for nBOCS-Random~(MAP), it is already quite large even at $t = 10^2$, and gradually converges to $R(t) \simeq 0.8$.
Although $R(t)$ converges at $t \simeq 7 \times 10^3$ in both cases, the value of $R(t)$ is always larger in nBOCS-Random (MAP) than in nBOCS (TS), and therefore MAP version is more likely to find the ground state earlier. This could cause different scaling of $[\tau]$.
Note that, regarding that $\vec w_J$ is dense with size $P = O(N^2)$, we expect that $O(N^2)$ data points are needed to find $\vec w_J$ in general, yielding the optimal exponent $z = 2$. Our algorithms, especially nBOCS-Random~(MAP), are thus very close to the optimal.

To conclude, we have studied the effects of postprocessing on Bayesian optimization for an NP-hard, high-dimensional optimization problem.
We have focused on the ground-state search of the SK model and found that, when combined with the nBOCS~(MAP) algorithm, the postprocessing drastically reduces the number of steps $[\tau]$ to find the ground state. 
We then found that $[\tau]$ has a power-law scaling with exponent $z$, which is slightly smaller for nBOCS-Random~(MAP) than the other algorithms.

Our results show that the conceptually simple, random postprocessing drives $\vec w$ to escape from a metastable, local optimum, yielding enhanced parameter space exploration.
We thus expect that it should improve Bayesian optimization for general optimization problems as well as the ground-state search of the SK model. 
Nevertheless, applying the postprocessing to Bayesian optimization of other problems, e.g., constrained combinatorial optimization problems, is certainly an interesting direction.

\begin{acknowledgement}
We thank Renichiro~Haba for useful discussions.
Y.~N. acknowledges support from JSPS KAKENHI (Grant No. 22K13968).
M.~O. receives financial support from JSPS KAKENHI Grant No. 23H01432, the MEXT-Quantum Leap Flagship Program Grant No. JPMXS0120352009, as well as Public/Private R\&D Investment Strategic Expansion PrograM (PRISM) and programs for Bridging the gap between R\&D and the IDeal society (society 5.0) and Generating Economic and social value (BRIDGE) from Cabinet Office.
\end{acknowledgement}

\bibliographystyle{jpsj}
\bibliography{references}

\begin{thebibliography}{10}

\bibitem{yamashita2018crystal}
T.~Yamashita, N.~Sato, H.~Kino, T.~Miyake, K.~Tsuda, and T.~Oguchi: Phys. Rev.
  Mater. {\bfseries 2} (2018) 013803.

\bibitem{ju2017designing}
S.~Ju, T.~Shiga, L.~Feng, Z.~Hou, K.~Tsuda, and J.~Shiomi: Phys. Rev. X
  {\bfseries 7} (2017) 021024.

\bibitem{kitai2020designing}
K.~Kitai, J.~Guo, S.~Ju, S.~Tanaka, K.~Tsuda, J.~Shiomi, and R.~Tamura: Phys.
  Rev. Res {\bfseries 2} (2020) 013319.

\bibitem{homma2020optimization}
K.~Homma, Y.~Liu, M.~Sumita, R.~Tamura, N.~Fushimi, J.~Iwata, K.~Tsuda, and
  C.~Kaneta: J. Phys. Chem. C {\bfseries 124} (2020) 12865.

\bibitem{Tanaka2023}
T.~Tanaka, M.~Sako, M.~Chiba, C.~Lee, H.~Cha, and M.~Ohzeki: J. Phys. Soc. Jpn.
  {\bfseries 92} (2023) 023001.

\bibitem{matsumori2022application}
T.~Matsumori, M.~Taki, and T.~Kadowaki: Sci. Rep. {\bfseries 12} (2022) 12143.

\bibitem{tibaldi2022bayesian}
S.~Tibaldi, D.~Vodola, E.~Tignone, and E.~Ercolessi: arXiv:2209.03824 .

\bibitem{tamura2020data}
R.~Tamura, K.~Hukushima, A.~Matsuo, K.~Kindo, and M.~Hase: Phys. Rev. B
  {\bfseries 101} (2020) 224435.

\bibitem{seki2022black}
Y.~Seki, R.~Tamura, and S.~Tanaka: arXiv:2209.01016 .

\bibitem{jones1998efficient}
D.~R. Jones, M.~Schonlau, and W.~J. Welch: J. Global Optim. {\bfseries 13}
  (1998) 455.

\bibitem{pelikan1999boa}
M.~Pelikan, D.~E. Goldberg, and E.~Cant{\'u}-Paz: Proc. GECCO-99, Vol.~1, 1999.

\bibitem{brochu2010tutorial}
E.~Brochu, V.~M. Cora, and N.~De~Freitas: arXiv:1012.2599 .

\bibitem{shahriari2015taking}
B.~Shahriari, K.~Swersky, Z.~Wang, R.~P. Adams, and N.~De~Freitas: Proc. IEEE
  {\bfseries 104} (2015) 148.

\bibitem{garnett_bayesoptbook_2023}
R.~Garnett: {\em {Bayesian Optimization}} (Cambridge University Press, 2023).

\bibitem{baptista2018bayesian}
R.~Baptista and M.~Poloczek: ICML, 2018, pp. 462--471.

\bibitem{Takahashi2018}
C.~Takahashi, M.~Ohzeki, S.~Okada, M.~Terabe, S.~Taguchi, and K.~Tanaka: J.
  Phys. Soc. Jpn {\bfseries 87} (2018) 074001.

\bibitem{oh2019combinatorial}
C.~Oh, J.~Tomczak, E.~Gavves, and M.~Welling: Adv. NeurIPS {\bfseries 32}
  (2019).

\bibitem{deshwal2021mercer}
A.~Deshwal, S.~Belakaria, and J.~R. Doppa: Proc. AAAI Conf. Artif. Intell.
  {\bfseries 35} (2021) 7210.

\bibitem{Koshikawa2021}
A.~S. Koshikawa, M.~Ohzeki, T.~Kadowaki, and K.~Tanaka: J. Phys. Soc. Jpn.
  {\bfseries 90} (2021) 064001.

\bibitem{deshwal2022bayesian}
A.~Deshwal, S.~Belakaria, J.~R. Doppa, and D.~H. Kim: Proc. AAAI Conf. Artif.
  Intell. {\bfseries 36} (2022) 6515.

\bibitem{dadkhahi2022fourier}
H.~Dadkhahi, J.~Rios, K.~Shanmugam, and P.~Das: Proc. AAAI Conf. Artif. Intell.
  {\bfseries 36} (2022) 10156.

\bibitem{kim2020surrogate}
S.~H. Kim and F.~Boukouvala: Computers \& Chemical Engineering {\bfseries 140}
  (2020) 106847.

\bibitem{nusslein2023black}
J.~N{\"u}{\ss}lein, C.~Roch, T.~Gabor, J.~Stein, C.~Linnhoff-Popien, and
  S.~Feld: ICCS 2023, 2023, p.~48.

\bibitem{luong2019bayesian}
P.~Luong, S.~Gupta, D.~Nguyen, S.~Rana, and S.~Venkatesh: AI 2019: 32nd
  Australasian Joint Conf., 2019, p. 473.

\bibitem{kadowaki2022lossy}
T.~Kadowaki and M.~Ambai: Sci. Rep. {\bfseries 12} (2022) 15482.

\bibitem{rendle2010factorization}
S.~Rendle: ICDM 2010, 2010, p. 995.

\bibitem{rendle2012factorization}
S.~Rendle: ACM Trans. Intell. Syst. Technol. {\bfseries 3} (2012) 1.

\bibitem{papalexopoulos2022constrained}
T.~P. Papalexopoulos, C.~Tjandraatmadja, R.~Anderson, J.~P. Vielma, and
  D.~Belanger: ICML, 2022, p. 17295.

\bibitem{sherrington1975solvable}
D.~Sherrington and S.~Kirkpatrick: Phys. Rev. Lett. {\bfseries 35} (1975) 1792.

\bibitem{carvalho2009handling}
C.~M. Carvalho, N.~G. Polson, and J.~G. Scott: Intern. Conf. Artif. Intell.
  Stat., 2009, p.~73.

\bibitem{carvalho2010horseshoe}
C.~M. Carvalho, N.~G. Polson, and J.~G. Scott: Biometrika {\bfseries 97} (2010)
  465.

\bibitem{makalic2015simple}
E.~Makalic and D.~F. Schmidt: IEEE Signal Process. Lett. {\bfseries 23} (2015)
  179.

\bibitem{bhadra2019lasso}
A.~Bhadra, J.~Datta, N.~G. Polson, and B.~Willard: Stat. Sci. {\bfseries 34}
  (2019) 405.

\bibitem{thompson1933likelihood}
W.~R. Thompson: Biometrika {\bfseries 25} (1933) 285.

\bibitem{thompson1935criterion}
W.~R. Thompson: Ann. Math. Stat. {\bfseries 6} (1935) 214.

\bibitem{agrawal2013thompson}
S.~Agrawal and N.~Goyal: ICML, 2013, pp. 127--135.

\bibitem{kirkpatrick1983optimization}
S.~Kirkpatrick, C.~D. Gelatt~Jr, and M.~P. Vecchi: Science {\bfseries 220}
  (1983) 671.

\bibitem{vcerny1985thermodynamical}
V.~{\v{C}}ern{\`y}: J. Optim. Theory Appl. {\bfseries 45} (1985) 41.

\bibitem{bertsimas1993simulated}
D.~Bertsimas and J.~Tsitsiklis: Stat. Sci. {\bfseries 8} (1993) 10.

\bibitem{arora2005non}
S.~Arora, E.~Berger, H.~Elad, G.~Kindler, and M.~Safra: FOCS 2005, 2005, pp.
  206--215.

\bibitem{de1995exact}
C.~De~Simone, M.~Diehl, M.~J{\"u}nger, P.~Mutzel, G.~Reinelt, and G.~Rinaldi:
  J. Stat. Phys. {\bfseries 80} (1995) 487.

\bibitem{mitchell2002branch}
J.~E. Mitchell: Handbook of applied optimization {\bfseries 1} (2002) 65.

\bibitem{gurobi}
{Gurobi Optimization, LLC}.
\newblock {Gurobi Optimizer Reference Manual}, 2023.

\bibitem{montanari2021optimization}
A.~Montanari: SIAM J. Comput. {\bfseries 0} (2021) FOCS19.

\end{thebibliography}

\end{document}